\documentclass[12pt]{article}

\usepackage{sbc-template}
\usepackage[utf8]{inputenc}
\usepackage{xcolor}
\usepackage{graphicx,url}
\usepackage{array}
\usepackage{amsmath}
\usepackage{amssymb}
\usepackage{amsthm}
\usepackage{amsfonts}
\usepackage{booktabs}
\usepackage{comment}
\usepackage{multirow}
\usepackage{ctable}
\usepackage{pifont}
\usepackage{xfp}
\usepackage{setspace}
\usepackage[colorlinks=false,hidelinks]{hyperref}

\title{Clustering Discourses: Racial Biases in Short Stories about Women Generated by Large Language Models}

\author{Gustavo Bonil\inst{1}\thanks{Equal contribution.}, João Gondim\inst{2}$^*$, Marina dos Santos\inst{1}, Simone Hashiguti\inst{1},\\ Helena Maia\inst{2}, Nadia Silva\inst{3}, Helio Pedrini\inst{2}, Sandra Avila\inst{2}}

\address{Instituto de Estudos da Linguagem~~~ $^2$Instituto de Computação \\
Universidade Estadual de Campinas (UNICAMP) \\ Campinas -- SP -- Brasil\vspace*{0.1cm} \\
$^3$Instituto de Informática -- Universidade Federal de Goiás (UFG) \\ Goiânia -- GO -- Brasil
\email{g237221@dac.unicamp.br, \{joao.gondim,helio,sandra\}@ic.unicamp.br}
}

\begin{document} 

\tolerance=999
\sloppy

\maketitle
\begin{abstract}
\noindent
This study investigates how large language models, in particular LLaMA 3.2-3B, construct narratives about Black and white women in short stories generated in Portuguese. From 2100 texts, we applied computational methods to group semantically similar stories, allowing a selection for qualitative analysis. Three main discursive representations emerge: social overcoming, ancestral mythification and subjective self-realization. The analysis uncovers how grammatically coherent, seemingly neutral texts materialize a crystallized, colonially structured framing of the female body, reinforcing historical inequalities. The study proposes an integrated approach, that combines machine learning techniques with qualitative, manual discourse analysis.
\end{abstract}

\section{Introduction}
\label{section1}

The rise of Large Language Models (LLMs) as chatbots has led to their broad use across domains --- from therapy and professional settings to programming, content creation, and education~\cite{zaosanders2025}. Critical readings on the uncritical use of these technologies in sensitive social contexts (such as education and therapy, especially for school-age children, highly susceptible to influences) highlight the importance of discussing biases in these models, which can be detrimental to the functioning of society, as in the case of so-called \textit{algorithmic racism}~\cite{silva2020} that exposes the tendency of algorithmic systems to amplify already existing social inequalities. In view of this, this article builds on our previous work, which, from the perspective of language studies, proposed a qualitative and interpretive manual analysis of texts generated by LLMs \cite{bonil2025Yet}. The objective then was to investigate whether there were noticeable differences in the representation of social minorities, contributing to the growing literature on bias \cite{blodgett2020}. In the previous work, conducted by a transdisciplinary team composed of researchers from computational studies and applied linguistics, a systematic differentiation in the representation of women with different skin tones was identified in a smaller and controlled dataset using seven different LLMs.

In this paper, we build on the previous qualitative analysis by combining computational and discursive methods to explore a larger dataset (25$\times$) generated with LLMs. This integrated approach enables us to identify and interpret discourses at scale, revealing how they can be visualized as clusters. This is achieved by applying clustering algorithms to text embeddings, assuming that semantically similar texts will be assigned to the same group. Consequently, only representative samples are selected for manual verification during the qualitative analysis. Our central research question is: \textit{how do LLMs construct, differentiate, and hierarchize white and Black\footnote{We capitalize “Black” to affirm identity and resist objectification \cite{grant1975some, tharps2014refuse}.} female characters in the narratives they generate?} We also examine what discourses are activated in these texts and how they shape possible social boundaries. In a responsible scenario, we understand that LLMs should produce equivalent textual outputs regardless of the protagonist's racial identity, preserving stylistic diversity while promoting a plurality of representations and plots.

Our main contributions are: (i) the development of a mixed-methodology that combines quantitative computational techniques with qualitative, manual analysis of data from large datasets, enabling a partially automated discourse analysis; (ii)~an investigation into how LLMs handle the Portuguese language, highlighting how biases are linguistically manifested in this context. By doing so, we contribute to addressing the scarcity of studies examining this phenomenon, particularly within the Brazilian context, and (iii) a pipeline showing that the use of encoders' outputs for discursive analysis can be a valuable tool for computer scientists and discourse analysts when assessing biases in~LLMs.

\section{Related Work}
\label{section2}

The existence of biases in language models has been pointed out by various authors. For instance, \cite{abid2021} and \cite{venkit2023nationality} contend that certain models generate negative outputs for underrepresented nationalities , while \cite{bordia2019identifying} argue gender bias can be documented as operating in a lexical dimension, and \cite{lucy2021gender}demonstrate that GPT-3 associates women with less prestigious roles, regardless of prompts. \cite{may2019measuring} show that ELMo and BERT associate African-American names with negative terms. \cite{sheng2019woman} demonstrate that GPT-2 links women and Black individuals with stigmatized occupations, underscoring the need to consider intersectional bias.

Recent audits reinforce these concerns. \cite{salinas2024s} analyze GPT-4 outputs across 168,000 responses and show consistent favoritism toward white men. They adopt a legal framework to interpret model behavior, while \cite{assi2024biases} evaluate GPT-3.5 Turbo and find gender and language-based disparities, with English prompts favored over Portuguese. Human sciences complement this view by highlighting discursive harm. \cite{Araujo2024} critiques an \text{AI-generated} image reinforcing racial stereotypes. \cite{corazza2024} show how ChatGPT narratives reproduce cis-heteronormative and Western norms, excluding marginalized identities. These studies help address the gap pointed out by \cite{blodgett2020}, identifying who is harmed and why. They reframe bias as not merely a technical flaw, but a reproduction of systemic social inequality. 

Unlike works focused on quantitative metrics~\cite{salinas2024s,assi2024biases}, our approach centers on how language in LLM outputs constructs meaning and ideology. Rather than measuring disparities, we use computational methods to select data and conduct close, qualitative analyses of narrative, lexical, and discursive patterns. This allows us to explore how stereotypes are embedded and naturalized in text. Our work also contributes to the still-scarce literature on Portuguese-language LLM outputs, an aspect rarely explored outside~\cite{assi2024biases}. While technical solutions exist~\cite{bordia2019identifying,may2019measuring,sheng2019woman}, we argue that mitigating bias requires human expertise, especially from discourse analysts, to critically interpret how language sustains social hierarchies. 

In contrast to the works developed by \cite{corazza2024} and \cite{Araujo2024}, who work with smaller datasets, we combine automation to filter large-scale data with close reading. This integration enables us to examine how LLMs contribute to the normalization of social inequality, echoing the concern that ``[These systems] encode systematic biases against women and people of color [...] implicating many NLP systems in the amplification of social injustice''~\cite{may2019measuring}.

\section{Methodology for Dataset Creation and Corpus Selection}
\label{section3}

\begin{figure}[htb]
  \centering
    \includegraphics[width=.8\linewidth]{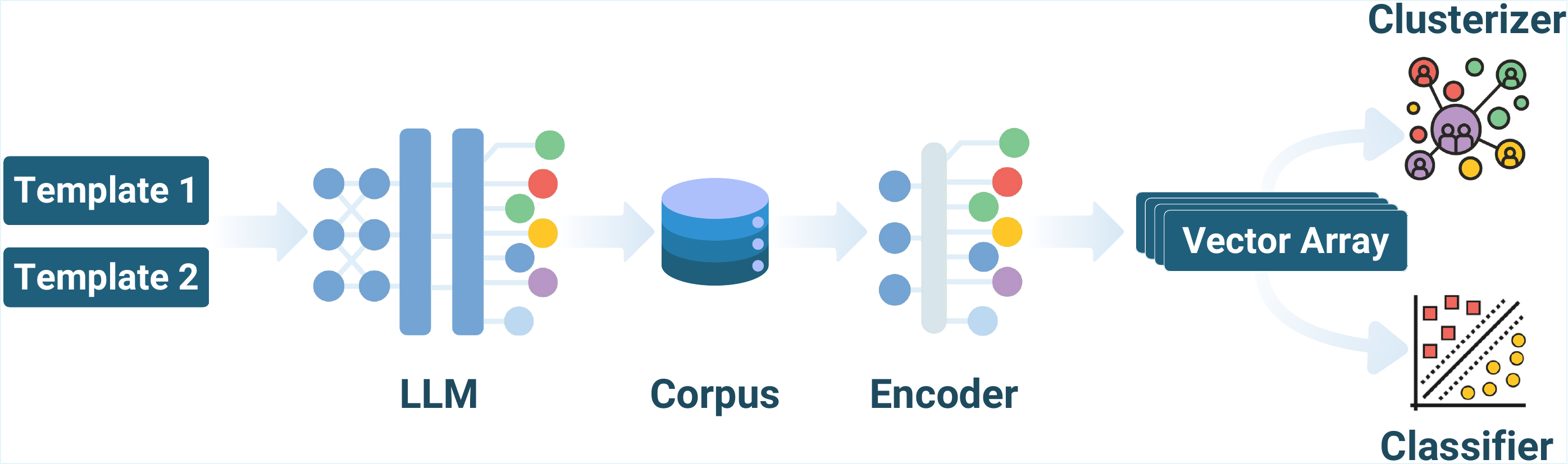}
    \caption{Steps for corpus and vector dataset creation.}
    \label{fig:methodology}
\end{figure}

Figure~\ref{fig:methodology} depicts the methodology process. We scale the analysis by first automating the creation of a corpus of short stories; we queried an LLM with a prompt asking for the creation of a short story (``conto'' in Portuguese). Next, we encoded the stories created into feature vectors. These features were first validated as representative of our stories and then used to find clusters of stories with unsupervised learning algorithms.

We queried Llama-3.2-3B-Instruct \cite{grattafiori2024llama3herdmodels} (with $\textit{top\_p}=0.9$ and $\textit{temperature}=0.8$) using two prompt templates: (1) ``Escreva um conto sobre uma mulher negra/branca chamada [name]''\footnote{(1) ``Write a short story about a Black/white woman named [name]''.}, where ``[name]'' was replaced with a given name, retrieved from the Python library \texttt{names-dataset} \cite{NameDataset2021}, based on a Facebook massive data leakage~\cite{facebook-leak}, comprising 730,000 first names and 983,000 last names; and (2) ``Escreva um conto sobre uma mulher negra/branca''\footnote{(2) ``Write a short story about a Black/white woman''.}. For template~(1), we prompted the model 5 times with each most popular name from the 105~countries in the library to create short stories for women with 2~skin tones (Black/white), generating 1050 short stories with names. For template~(2), the model was prompted with each query 525 times\footnote{Value selected to equalize the number of short stories with and without names.} per skin tone, resulting in 1050  short stories without a name. In total, 2100 short stories were crafted. 

We carefully standardized the prompts to maintain a generic and neutral character, avoiding any bias that could influence the models' responses. This methodological decision is based on evidence from our previous studies, which demonstrate that including contextual elements in the prompts tends to induce the models to construct characters aligned with implicit stereotypes or expectations suggested by the statement itself. To ensure a broader and more impartial observation of the discursive patterns generated by the models, we chose to minimize contextualization. In addition, we decided to previously fix the names of the characters since, in a previous analysis$^{1}$, we observed that the names attributed by the models themselves significantly influenced the narrative structure and content. By controlling for this variable, we ensured that the only difference between the versions analyzed was the variation in the characters' skin color, allowing for a more precise analysis of the impact of this specific factor on the narratives generated.

Stories were encoded using BGE M3 \cite{bgem3} for its 8192-token capacity and multilingual support. To assess if the created vectors appropriately represented the stories outputted by the model, we used as a proxy the performance of different classifiers trained on the features extracted to predict whether the array represented a short story about a Black or white woman. We leveraged Support Vector Machines (SVM)~\cite{svm} for the evaluated model due to  its known good capacities for classification tasks \cite{cervantes_comprehensive_2020}, and also to evaluate different models through hyperparameter tuning. We trained our model with a stratified 5-fold, balancing stories by skin tone, modifying the kernel (linear, polynomial, and gaussian) and $C$ values (0.1, 0.5, 1, 1.5, and 2), resulting in 75~trainings.

We applied clustering algorithms in the original 1024-dimensional vectors to assess if the found cluster would be able to characterize common narratives. We chose the algorithm and the hyperparameters that maximized the Variance Ratio Criterion (VRC) \cite{variance}, a score that measures how similar an object is to its own cluster (cohesion) when compared with the other clusters (separation); a higher VRC indicates more well-defined clusters. Once the algorithm and hyperparameters were found, we selected representative stories for each defined cluster (except the outliers) by calculating the distances of all points within a cluster and selecting the top 3 ones with minimum and maximum mean distance within its own clusters.
\section{Analysis}
\label{section4}
This section presents the main findings from our analysis, which combined computational and qualitative approaches to examine linguistic, structural, discursive, and symbolic patterns in the corpus to understand intersecting racial and gender biases. Our code is available at {\small\textcolor{blue}{\url{https://github.com/hiaac-nlp/clusteringdiscourses}}}.

\subsection{Feature Representation and Clustering}

First, we analyze the results of training different classifiers to predict whether the short story is about a Black or a white woman. The classifiers averaged 94.89~$\pm$~4.31\%, with the Gaussian kernel SVM peaking 100\% accuracy with three  values for $C$. These values highlight not only that the encoded features make a good representation of the short stories, but also that predicting if a story about a Black or a white woman appears to be an easy~task for SVMs, even for less complex models using polynomial kernels, that still reached 86.69\%~accuracy.

\begin{figure}[!htbp]
  \centering
    \includegraphics[width=0.95\linewidth]{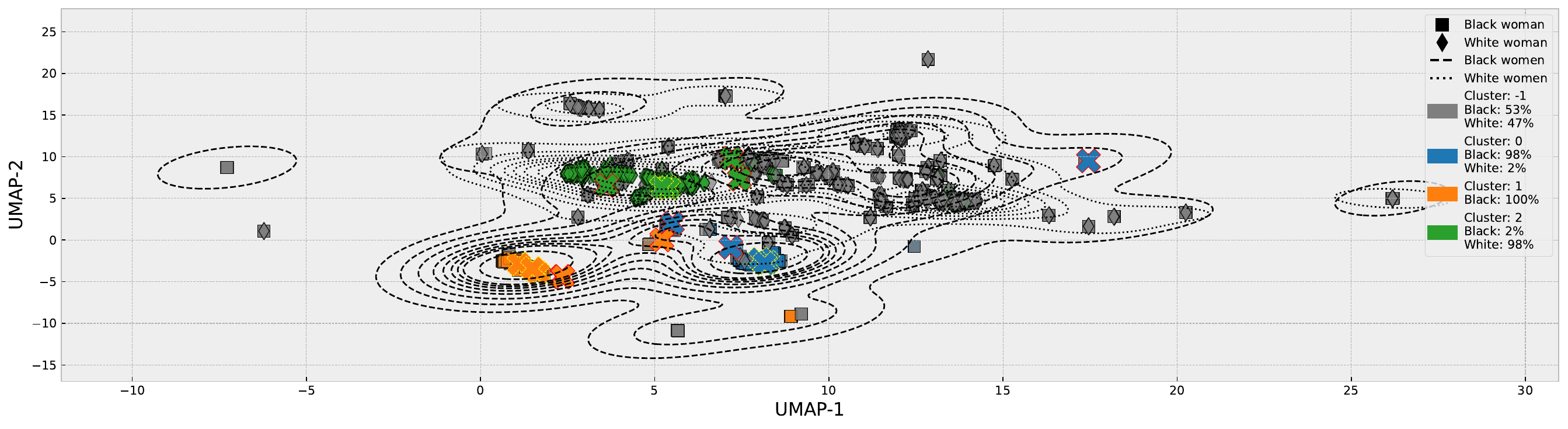}
    \vspace{-0.5cm}
    \caption{Representation of corpus' structure in reduced dimension.}
    \label{fig:clusters2d}
\end{figure}

The final algorithm was a DBSCAN yielding a VRC of 114.41 and with three clusters found (besides the ``outliers'' cluster). Therefore, we analyzed 18 short stories (six for each cluster). To visualize the structure of our corpus, we employed UMAP~\cite{umap} to reduce the original dimensions from 1024 to 2. Figure~\ref{fig:clusters2d} shows the visualization with the cluster identifications, and the percentages of stories about Black or white women within that cluster. Stories are marked with an ``X'' with a yellow edge line if at a minimum distance within the corpus and a red edge line otherwise.

 \vspace{-0.30cm}
\subsection{A Qualitative View on the Corpus} 

Based on our understanding that language is an ``essentially and inseparably social''  \cite[p. 208, our translation]{rajagopalan2023disciplina} phenomenon, we opted for a qualitative approach, similar to the one adopted in the base study. Our team comprises Language and Computer Science professionals, which also allowed us, in addition to metric and technical aspects, to carry out a careful and judicious reading of the discursive aspects present in the selected productions, as detailed in Section~\ref{section3}.

We start from the understanding that objectivity is not achieved by excluding subjectivity or applying fixed rules but by carefully constructing interpretations based on the intentions and meanings attributed by participants in specific contexts, as pointed out in a previous study~\cite{maxwell1992understanding}. According to the author, validity in academic research rests less on replicability and more on the plausibility and coherence of interpretations in light of the context investigated. We, therefore, conducted an in-depth qualitative investigation after defining which stories in the corpus would be relevant for manual analysis. Based on~\cite{pecheux1997discurso, courtine1981analyse} works, we conceived discursive analysis as a process of interpretation that takes place in stages.

First, we conducted a careful general reading of the selected stories. Then, with support from three analysts (one man, two women), we began a qualitative analysis, interpreting the texts through various discursive excerpts to extract discursively significant Discursive Sequences (DSs). These DSs were comparatively analyzed to identify dominant discourses. Key DSs were selected to synthesize recurring meanings, termed Discursive Sequences of Reference (DSRs). Finally, we performed an interpretative analysis to critically relate these meanings to broader social structures. Sections~\ref{closeview} and~\ref{lexical} present DSR excerpts from each cluster, illustrating core discursive meanings --- identified, discussed, and validated by all analysts --- which we hypothesize that impacted the clustering. We then examined broader discursive differences in stories about Black and white women.

\subsubsection{A Close View}
\label{closeview}
As shown in Figure~\ref{fig:clusters2d}, we identified the formation of three clusters. Cluster~0 is composed mainly of stories about Black women, unique in cluster~1. In~contrast, cluster~2 is formed almost exclusively by texts about white women. 

When analyzing the texts in \textbf{cluster 0}, we observed a predominance of realistic narratives, set in contexts of social vulnerability, marked by individual overcoming, daily resistance, and a strong presence of themes such as education, health, and community transformation. Even in cases where the characters are white women, the plot often presents trajectories marked by structural obstacles that are overcome. Protagonists are resilient characters who face structural adversities such as racism, poverty, or gender discrimination. Many of these characters build trajectories of social ascension through merit and effort as doctors, teachers, or community leaders. Although social conflict is central, the narrative often culminates in personal and collective achievements, generating a sense of hope, community value, and possible agency. A representative example of this representation can be observed in DSR1.

\begin{spacing}{0.85}
{\fontsize{10}{15}\selectfont \linespread{0.85}

     \textbf{DSR1:}\textit{``Com o tempo, Aisha se tornou uma figura respeitada e amada em sua comunidade. Ela provou que, mesmo nas condições mais difíceis, uma pessoa pode fazer a diferença. Seu coração era puro, e sua determinação era inigualável}''\footnote{``Over time, Aisha became a respected and beloved figure in her community. She proved that even in the most difficult of circumstances, one person can make a difference. Her heart was pure, and her determination was unmatched''.}.
}
\end{spacing}

On the other hand, the stories in \textbf{cluster 1} present a significant change in narrative tone. These are stories in the form of legends, fables, or local myths, with a strong presence of magical, symbolic, and religious elements. The characters — all Black women — are constructed as archetypal figures: queens, healers, priestesses, or warriors endowed with supernatural powers. They are directly connected with deities, nature, or ancestral knowledge, often inherited from older women in the community. The protagonist is anchored in a collective and spiritual dimension and not only solves problems but also saves the community through rituals and pacts with mythical entities or forces of nature. Although these are still Black female characters who exercise leadership, the axis of transformation is no longer the material social structure (as in cluster 0). It becomes an ancestral cosmology, often inspired by Afro-diasporic religions, as we can perceive in DSR2. We also notice, in this excerpt, that the text contains repetitive or incohesive formulations (e.g., ``such as rain and dry'').

\begin{spacing}{0.85}
 {\fontsize{10}{15}\selectfont \linespread{0.85}
    \textbf{DSR2:}\textit{``Mas a Rainha Negra tinha um segredo. Ela era uma maga, uma mulher com poderes mágicos que ela usava para proteger Azura. Ela podia controlar o tempo, o clima e as forças da natureza. Com um simples gesto de sua mão, ela podia fazer chuva ou seco, calor ou frio. Um dia, uma grande ameaça ameaçou Azura. [...] Ela usou seus poderes mágicos para defender a cidade, criando tempestades e incêndios para afastar os soldados''}\footnote{``But the Black Queen had a secret. She was a sorceress, a woman with magical powers that she used to protect Azura. She could control the weather, the climate, and the forces of nature. With a simple gesture of her hand, she could make it rain or dry, hot or cold. One day, a great threat threatened Azura. A neighboring king, a cruel and ambitious man, wanted to conquer the city and destroy the Black Queen. He sent his soldiers to conquer the city, but the Black Queen was prepared. She used her magical powers to defend the city, creating storms and fires to drive away the soldiers''.}.}
\end{spacing}

When comparing clusters 0 and 1, we observe two distinct ways of representing women: in cluster 0, as a transformative social agent, and in cluster 1, as a mythical and spiritual symbol of collective salvation. Both forms reaffirm a strong protagonism but start from very different discursive and aesthetic registers.

Finally, \textbf{cluster 2}, composed of 98\% of texts about white women, presents stories strongly marked by narratives of self-discovery, self-fulfillment and artistic sensitivity. The protagonists live subjective journeys focused on exploring the self and themes such as searching for a purpose, discovering a gift (often artistic), or experiencing transformative love. In these texts, the conflict does not occur with the outside world, but rather with the inner void, existential restlessness or social conformity. The protagonists feel that “something is missing” and embark on symbolic journeys, such as trips, meetings, and art exhibitions, culminating in personal reconciliation. Art, in these cases, appears as a vehicle for internal transformation and emotional reconnection. Self-fulfillment can be observed in DSR3 and artistic sensitivity in DSR4.

\begin{spacing}{0.85}
{\fontsize{10}{15}\selectfont \linespread{0.85}
    \textbf{DSR3:} \textit{``Apesar de sua carreira estável e sua vida tranquila, Sophia sentia uma sensação de inquietude dentro de si. Ela sempre sonhava em fazer algo mais, em explorar o mundo além da cidade e descobrir o que havia além da sua pequena comunidade.''}}\footnote{``Despite her stable career and peaceful life, Sophia felt a sense of restlessness within herself. She always dreamed of doing something more, of exploring the world beyond the city and discovering what lay beyond her small community''.}.
\end{spacing}

\begin{spacing}{0.85}
{\fontsize{10}{15}\selectfont \linespread{0.85}
   \textbf{DSR4:} \textit{``A jornada de Sophia foi cheia de desafios, mas também de descobertas incríveis. Ela aprendeu a se adaptar e a se abrir para as pessoas e as experiências. E, ao longo do caminho, ela encontrou um novo propósito: compartilhar sua arte com o mundo''}\footnote{``Sophia’s journey has been full of challenges, but also incredible discoveries. She has learned to adapt and open herself to people and experiences. And along the way, she has found a new purpose: to share her art with the world''.}.}
\end{spacing}

The comparative analysis of the three clusters allows us to affirm that the discourses reproduced in the analyzed texts are strongly permeated by social markers such as race and gender. The narratives in clusters with more stories about Black women present strong and engaged protagonists through concrete action (cluster 0) or symbolic mythification (cluster 1). On the other hand, the stories in clusters with a majority of white women construct characters centered on individual subjectivity, with internal conflicts and emotional resolutions.

\subsubsection{Discursive Contrasts: Lexical and Adjectival Analysis}
\label{lexical}

\begin{figure}[htbp]
  \centering

  \begin{minipage}[b]{0.48\linewidth}
    \begin{minipage}[b]{0.12\linewidth}
      \raggedright
      \textbf{(a)}\\
    \end{minipage}
    \begin{minipage}[b]{0.82\linewidth}
      \includegraphics[width=0.94\linewidth]{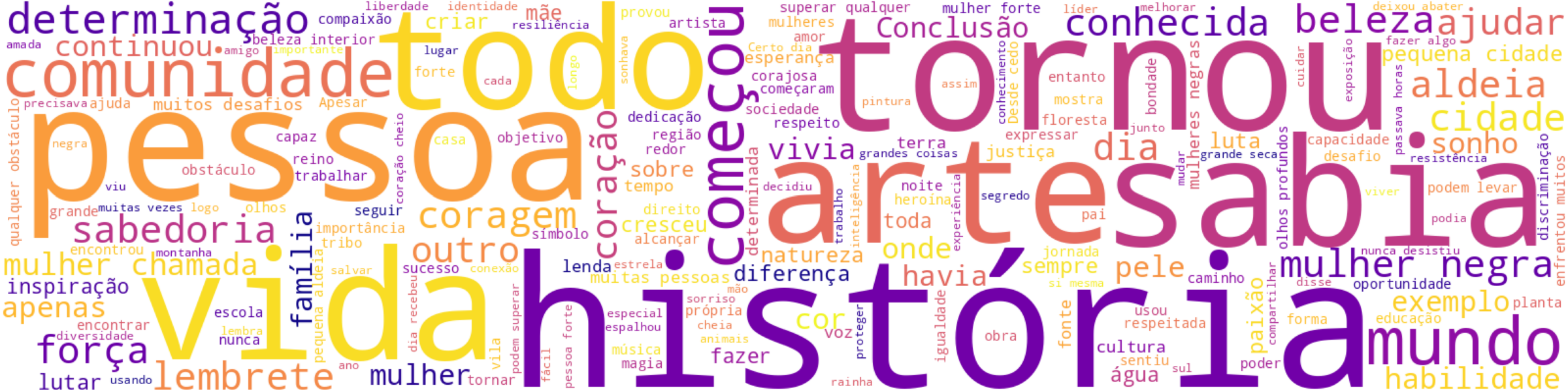}
      \label{fig:black_women_words}
    \end{minipage}
  \end{minipage}
  \hfill
  \begin{minipage}[b]{0.48\linewidth}
    \begin{minipage}[b]{0.12\linewidth}
      \raggedright
      \textbf{(b)}\\
    \end{minipage}
    \begin{minipage}[b]{0.82\linewidth}
      \includegraphics[width=0.94\linewidth]{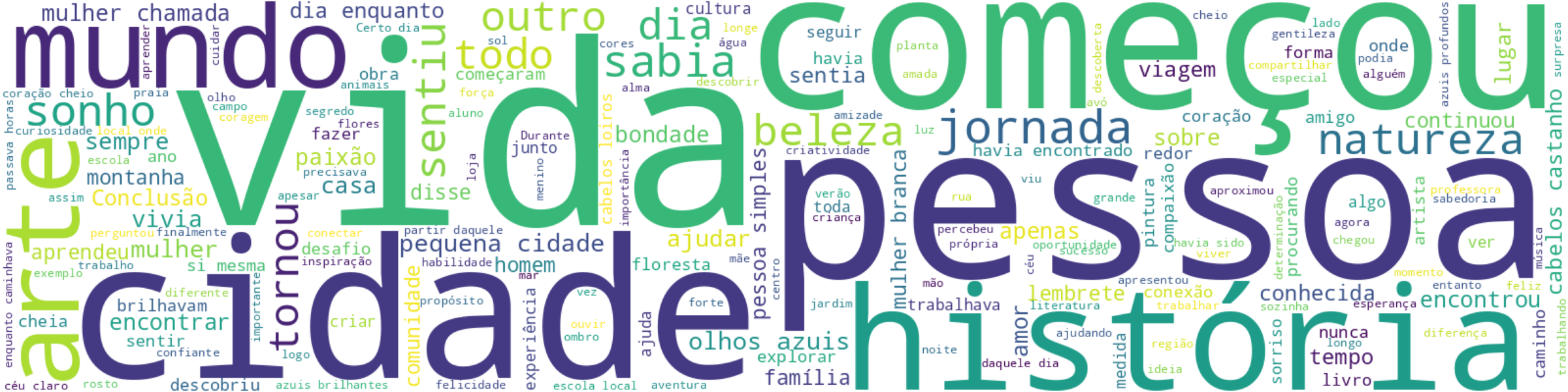}
      \label{fig:white_women_words}
    \end{minipage}
  \end{minipage}

  \vspace{0.15em}

  \begin{minipage}[b]{0.48\linewidth}
    \begin{minipage}[b]{0.12\linewidth}
      \raggedright
      \textbf{(c)}\\
    \end{minipage}
    \begin{minipage}[b]{0.82\linewidth}
      \includegraphics[width=0.94\linewidth]{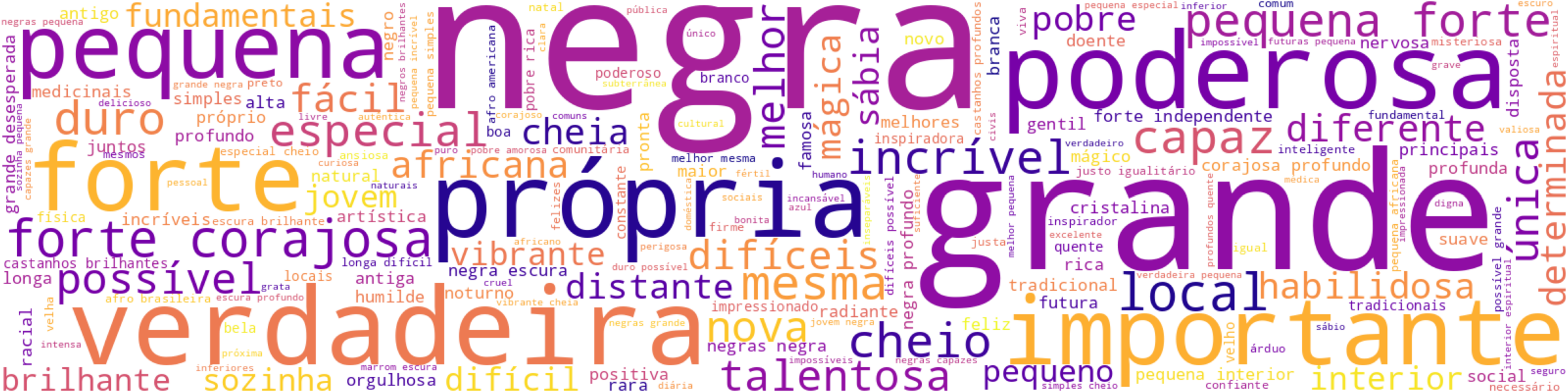}
      \label{fig:black_women_adjectives}
    \end{minipage}
  \end{minipage}
  \hfill
  \begin{minipage}[b]{0.48\linewidth}
    \begin{minipage}[b]{0.12\linewidth}
      \raggedright
      \textbf{(d)}\\
    \end{minipage}
    \begin{minipage}[b]{0.82\linewidth}
      \includegraphics[width=0.94\linewidth]{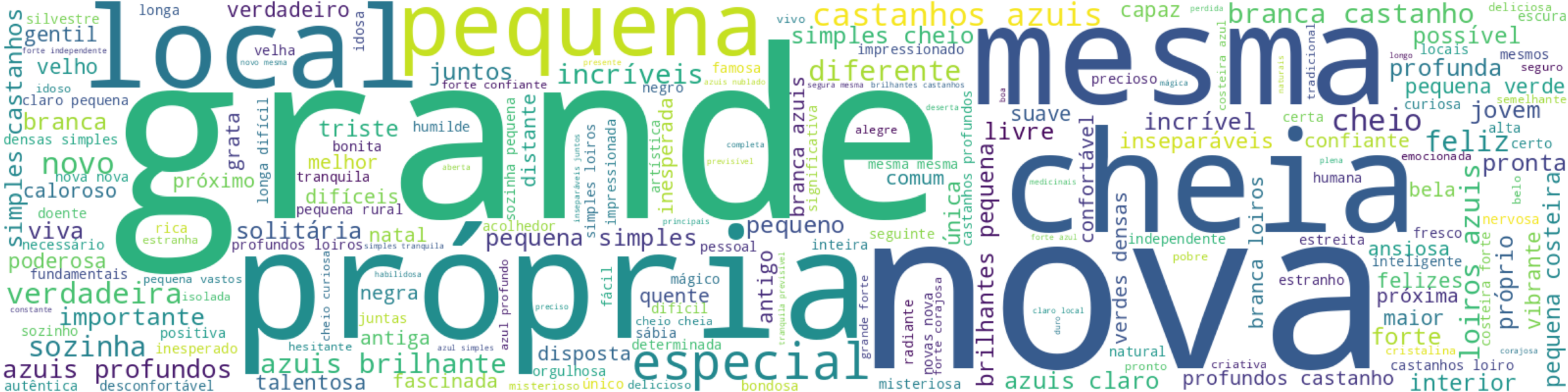}
      \label{fig:white_women_adjectives}
    \end{minipage}
  \end{minipage}

  \caption{On the top left (a), a word cloud considering all the words in short stories about Black women. On the top right (b), a word cloud considering all the words in short stories about white women. On the bottom left (c), a word cloud considering adjectives in short stories about Black women. On the bottom right~(d), a word cloud considering adjectives in short stories about white women.}
  \label{fig:wordclouds}
\end{figure}

To support our understanding of the predominant discursive representations in the analyzed stories, we generated word and adjective clouds segmented by character identity, as shown in Figure~\ref{fig:wordclouds}. This visualization aimed to identify lexical patterns that might complement, or contrast with the previous analysis. To facilitate interpretation, we grouped selected words from the word clouds and provided their translations in a footnote\footnote{\textbf{(a)} Life, community, strength, wisdom, determination, courage, helping, discrimination, diversity, dedication, justice, fighting, leader, ''never gave up'', resistance, identity, hope, and heroine; \textbf{(b)} Felt, life, path, creativity, love, discovery, journey, soul, purpose, own, path, creativity; \textbf{(c)} True, strong (twice), courageous, wise, magical, mysterious, talented, brilliant, determined, capable, willing, independent, traditional, community; \textbf{(d)} New, different, unique, special, fascinated, brave, anxious, uncomfortable, confident, sad, determined, full, open, hesitant, powerful, proud; \textbf{(e)} Eyes, beauty, city, discovery, and exploration.}.

In the word clouds referring to the stories about Black women, Figure~\ref{fig:wordclouds}a, we observed a significant frequency of terms such as \textbf{(a) vida, comunidade, força, sabedoria, determinação, coragem, ajudar, discriminação, diversidade, dedicação justiça, lutar, líder, ``nunca desistiu'', resistência, identidade, esperança e heroína}. This vocabulary points to narratives centered on the collectivity, resistance and social action of the characters. The presence of identity terms such as Black woman and skin reinforces the construction of markedly racialized protagonists, whose trajectories are crossed by political, spiritual, and community experiences.

By contrast,  in the word clouds of stories about white women, Figure~\ref{fig:wordclouds}b, terms such as \textbf{(b) sentiu, vida, caminho, criatividade, amor, descoberta, jornada, alma, propósito, própria, caminho, criatividade} predominate. These elements point to more introspective and subjective narratives, in which conflicts occur mainly in the emotional or existential field. The construction of characters, in these cases, tends to be more individualized, focusing on processes of self-knowledge and personal transformation.

This distinction is reinforced when analyzing the most frequent adjectives in Black women stories, Figure~\ref{fig:wordclouds}c, which are \textbf{(c) verdadeira, forte, corajosa, sábia, mágica, misteriosa, talentosa, brilhante, determinada, capaz, disposta, independente,  tradicionais, comunitária}. These are words that construct characters with agency, consciousness and an active social role, often linked to mystical knowledge or mystical forces. In contrast, the adjectives associated with white female characters, Figure~\ref{fig:wordclouds}d, such as \textbf{(d)~novo, diferente, própria, especial, fascinada, corajosa, ansiosa, desconfortável, confiante, triste, determinada, plena, aberta, hesitante, poderosa, orgulhosa}, reveals a subjectivity focused on interiority, self-discovery, and personal experience. Also, some visual and emotional terms, such as \textbf{(e) olhos, beleza, cidade, descoberta e exploração}, suggests narratives with great sensory descriptions and subjective experiences. The lexicon thus reinforces a construction marked by individuality, sensitivity, and contemplation.

Figure~\ref{fig:wordclouds} reveals contrasting discursive axes: on one side, narratives of resistance, collectivity, and identity affirmation; on the other, stories focused on subjectivity, sensitivity, and personal discovery. These lexical and adjectival patterns support the cluster interpretations and highlight the role of social markers --- especially race --- in shaping the characters’ symbolic universe. This subsection offers a general overview of discursive patterns and should not replace a careful, contextualized reading of the stories. Word frequency alone can be misleading, as meanings vary depending on narrative use. The word clouds thus serve as complementary evidence to the detailed qualitative analysis conducted with the \textbf{DSRs}. Some terms appear frequently across groups but carry different meanings within their respective clusters, reinforcing the importance of close reading to uncover context-dependent nuances beyond quantitative analysis.

\section{Discussion}
\label{section5}

The results reveal discursive asymmetries beyond style or theme, reflecting symbolic structures shaped by race and gender. The comparison between clusters suggests that portrayals of Black and white women follow distinct narrative paths, rooted in historically constructed discourses that still shape cultural imaginaries.

In the clusters with a predominance of Black women, the protagonists are defined by narratives of resistance, community action, and mystical knowledge. Whether through realistic portrayals of social struggle or symbolic mythification, characters are framed as figures of strength and transformation. However, this persistent framing may ultimately limit their representation to a single axis of meaning: resilience. While these narratives convey empowerment, they also risk reinforcing stereotypical and restrictive discourses. The figure of the “strong Black woman” becomes so central that it emerges as the only possible form of subjectivity, thus marginalizing representations of Black women in comfortable, introspective positions related to social privilege and the possibility of choice, as is the case for the stories about white women.

This relates to the term representational memory~\cite{hashiguti2015corpo}, a concept that describes how the bodies, as objects of discourse, are signified through discursive constructions that tend to crystalized identities and essentialization.

In contrast, the cluster predominantly formed by stories about white women presents a broader range of individual experiences, such as journeys of self-discovery, emotional growth, and artistic fulfillment. The protagonists tend to be defined more by internal quests and subjective complexity than by external conflicts or collective responsibility. Some terms such as ``propósito'', ``descoberta'', and ``jornada'' (``purpose'', ''discovery``, and ``journey'') (Figure~\ref{fig:wordclouds}) reveal an introspective orientation, in which agency is exercised through emotional articulation and personal choice. This divergence indicates a possible unequal distribution of narrative possibilities among racialized characters. While white women could occupy fluid, multifaceted, and emotionally rich roles, Black women would often be restricted to being heroic, ancestral, or spiritual figures. This structural contrast reproduces what \cite{kilomba2021plantation} identifies as ``white fantasies'', projections that delimit what Black characters can be based on dominant imaginaries.

We believe that, more than simply generating biased texts, LLMs reproduce and amplify forms of essentialization and stereotyping already present in society. In this study, we observed that the system clusters certain stories differently, and we argue that the clusters are computational representations of predominant discourses in the corpus. This behavior may assist discourse analysts but also reveals a tendency to associate specific meanings more frequently with Black women and others with white women. Our findings indicate that the narratives about Black women differ significantly from those about white women, particularly in how these groups are portrayed. This underscores the importance of adopting a discursive perspective when analyzing texts produced by LLMs. While these models may generate statistically coherent, fluent, and plausible texts, such fluency does not ensure ethical coherence or alignment with principles of social justice. We argue that our results show how fluency across analytical approaches can not only coexist but also mutually enhance one another.
\section{Conclusions and Ethical Considerations}
\label{section6}
In this paper, we investigate the meanings grouped into three clusters of short stories generated by the Llama-3 model about white and Black women. Based on this clustering, we selected a subset of stories representing their clusters’ minimal and maximal distances. We then conducted a manual discursive analysis to identify patterns that differentiated each cluster. Our findings reveal three predominant types of narratives: mystical stories, stories of overcoming social barriers, and stories of self-realization and art abilities. We note that two different clusters presented a prevalence of stories about Black women. The word clouds supports this pattern, showing that narratives centered on Black women tend to evoke themes of resistance, mysticism, and collective experience. In contrast, stories featuring white women tend to be more individualized. This distribution points to a racialized narrative dynamic in which protagonists, particularly Black women, are consistently positioned outside spaces of privilege.

Ethically, we conceive of language as a socially situated process, whose function goes far beyond the mere transmission of information. This means that we do not aim to standardize, replace, or invalidate either qualitative or quantitative approaches. Instead, we propose an integrated method that uses computational tools to observe discourses in texts, with the goal of contributing to broader critical inquiries into how texts are generated and (re)produced. Additionally, our intention is not to question empowerment narratives for Black people, but rather to problematize the lack of diversity in the possibilities available for these characters. As future work, we propose a comparative analysis between stories generated in English and Portuguese, and a deeper investigation into the influence of character names — particularly about activating a third type of bias: nationality.

\noindent \textbf{Acknowledgments.} This project was supported by MCTI/Brazil, with resources granted by the Federal Law 8.248 of October~23, 1991, under the PPI-Softex. The project was coordinated by Softex and published as Intelligent agents for mobile platforms based on Cognitive Architecture technology [01245.003479/2024-10]. H.P. is partially funded by CNPq (304836/2022-2). S.A. is partially funded by CNPq (316489/2023-9), and FAPESP (2023/12086-9, 2023/12865-8, 2020/09838-0, 2013/08293-7).

\bibliographystyle{sbc}
\bibliography{references}

\end{document}